\documentclass[conference]{IEEEtran}

\pdfoutput=1
\usepackage{color}
\usepackage{graphicx}
\usepackage{algorithmic}
\usepackage{algorithm}
\ifCLASSINFOpdf
\else
\fi
\begin{document}
%
\title{Biogeography-Based Informative Gene Selection and Cancer Classification Using SVM and Random Forests}

\author{\IEEEauthorblockN{Sarvesh Nikumbh}
\IEEEauthorblockA{Centre for Modeling and Simulation,\\University of Pune (UoP),\\Pune, India\\
sarvesh@cms.unipune.ac.in\\
}
\and
\IEEEauthorblockN{Shameek Ghosh and V. K. Jayaraman}
\IEEEauthorblockA{Evolutionary Computing and Image Processing (ECIP),\\
Center for Development of Advanced Computing(C-DAC),\\
Pune, India\\ \{shameekg, jayaramanv\}@cdac.in}
}

%


\maketitle

\begin{abstract}
 Microarray cancer gene expression data comprise of very high dimensions. Reducing the dimensions helps in improving the overall analysis and classification performance. We propose two hybrid techniques, Biogeography \textendash \space based Optimization \textendash \space Random Forests (BBO \textendash \space RF) and BBO \textendash \space SVM (Support Vector Machines) with gene ranking as a heuristic, for microarray gene expression analysis. This heuristic is obtained from information gain filter ranking procedure. The BBO algorithm generates a population of candidate subset of genes, as part of an ecosystem of habitats, and employs the migration and mutation processes across multiple generations of the population to improve the classification accuracy. The fitness of each gene subset is assessed by the classifiers \textendash \space SVM and Random Forests. The performances of these hybrid techniques are evaluated on three cancer gene expression datasets retrieved from the Kent Ridge Biomedical datasets collection and the libSVM data repository. Our results demonstrate that genes selected by the proposed techniques yield classification accuracies comparable to previously reported algorithms.

\end{abstract}


%
\IEEEpeerreviewmaketitle

\section{Introduction}
%
%

Microarray gene expression experiments help in the measurement of expression levels of thousands of genes simultaneously. Such data help in diagnosing various types of tumors with better accuracy. The fact that this process generates a lot of complex data happens to be its major limitation. Normally the number of genes (features) is much greater than the number of samples (instances) in a microarray gene expression dataset. Such structures pose problems to machine learning and make the problem of classification difficult to solve. This is mainly because, out of thousands of genes, most of the genes do not contribute to the classification process. As a result gene subset selection acquires extreme importance towards the construction of efficient classifiers with high predictive accuracy.

To overcome this problem, one way is to select a small subset of informative genes from the data. This technique which is known as \textit{gene selection} or \textit{feature selection} helps in tackling overfitting by getting rid of noisy genes,  reducing the computational load and in increasing the overall classification performance of the learning models.

Gene selection algorithms are mainly categorized as : wrappers and filters. Wrappers make use of learning algorithms to estimate the quality or suitability of genes to the modelling problem. Optimization algorithms in combination with various classifiers fall into this category as described in \cite{vkj-aco, aco-rf, vkj-book}. On the other hand, filters  \cite{john-feature} evaluate the genes considering their inherent characteristics without making use of a learning algorithm. Filters, therefore, give an insight into the properties of the dataset we use. Algorithms based on statistical tests and mutual information are some examples of filters.

This paper presents hybrid BBO \textendash \space RF and hybrid BBO \textendash \space SVM approaches for simultaneous informative gene selection and high performance classification. Additionally, for enhancing performance we provide information gain gene ranking as heuristic knowledge to our BBO algorithm. It traverses the enormously large search space by using this ranking information to iteratively obtain informative gene subsets. The selected subsets of genes (candidate solutions) in each generation are subsequently evaluated by SVM and Random Forests CV (cross validation) accuracies.

\section{Methodology}
\subsection{Biogeography\textendash based Optimization}
Biogeography is the study of geographical distribution of species over geological period of time. Biological literature on the same is massive. In 2008, for the first time, Simon \cite{simonbbo} applied the biogeography analogy to the idea of engineering optimization and thus introduced the Biogeography\textendash based Optimization (BBO) technique. It is a  population based method that works with a collection of candidate solutions over generations. It attempts to explore the  combinatorially large solution spaces with a stochastic approach like many other evolutionary algorithms \cite{pso-eberhart, ea-de}. It mimics the geographical distribution of species to represent the problem and its candidate solutions in the search space, subsequently using the process of species migration and mutation to redistribute solution instances across the search space in quest of globally optimal or near optimal solutions.

BBO, as is or in variations, has been explored for various combinatorial and constrained/unconstrained optimization problems \cite{bboblended} including the likes of the Traveling Salesman Problem \cite{bbotspMoXu,bbotspSong}, satellite image classification \cite{panchal} and sensor selection \cite{simonbbo} among others. But as of 2012, no work is reported of using BBO as a gene selection technique for microarray gene expression data analysis. We attempt to study BBO for gene selection and classification in this work.

In BBO, there exists an ecosystem (population) which in turn consists of a number of habitats (islands). Each habitat has a \textit{habitat suitability index} (HSI), which is similar to a fitness function and depends on many features/attributes  of the island. If a value is assigned to each feature, then the HSI of a habitat \textit{H} is a function of these values. These variables characterizing a habitat's suitability collectively form the \textit{\textquoteleft suitability index variables\textquoteright} \space (SIVs). Thus,
\begin{displaymath}
 HSI (Habitat_{i}) \space \rightarrow \space f(SIV_{1}, SIV_{2}, \ldots , SIV_{m})
\end{displaymath}

For the problem of gene selection, the SIVs of a habitat (candidate solution) are the selected subsets of genes out of the
set of all genes. The ecosystem is therefore a random collection of candidate gene subsets.

\begin{figure}[h]
\centering
\includegraphics[width=2.0in]{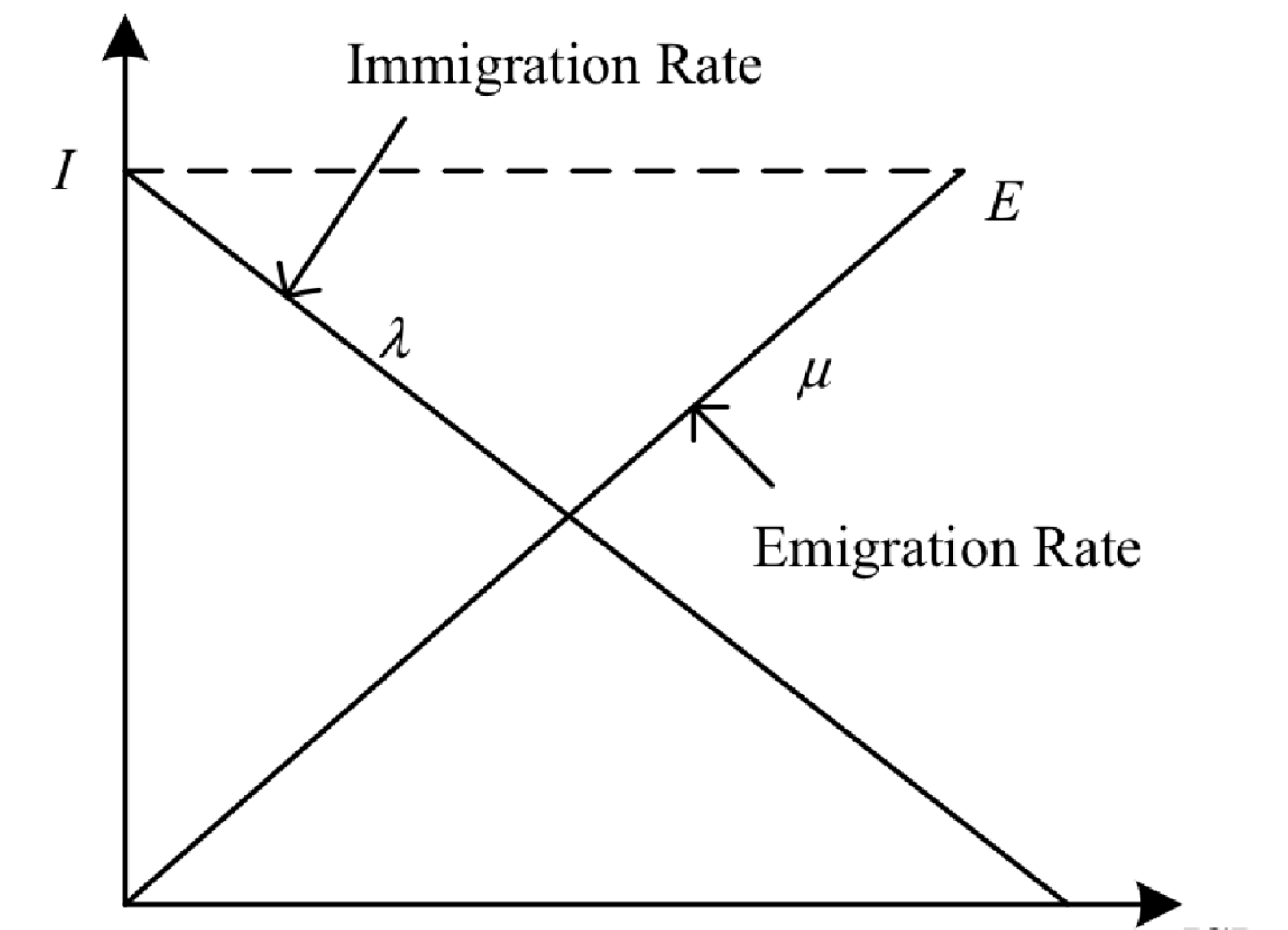}
\caption{Migration rate vs. No. of species}
\label{fig_sim}
\end{figure}

A good solution is thus analogous to a good HSI and \textit{vice versa}. Good HSI solutions tend to share SIVs with poor HSI solutions. This form of sharing, termed as migration, is controlled by emigration and immigration rates of the habitats. We have purposely kept the model simple and have obeyed the original simple linear model for migration as shown in Figure \ref{fig_sim}.

where E and I are the maximum emigration and immigration rates, both typically set to 1. The individual immigration and emigration rates ($\lambda$ and $\mu$ respectively) are calculated by the same formulae for this simple linear model as in \cite{simonbbo}.

\begin{displaymath}
 \lambda_{k} = I\left( 1 - \frac{k}{n}\right)
\end{displaymath}
\begin{displaymath}
 \mu_{k} = \frac{Ek}{n}
\end{displaymath}

where $k$ is the iterator for the $n$ habitats.


\subsection{BBO Gene Selection Algorithm}
We present our BBO algorithm for performing gene selection. 
For our problem, we treat a gene (identified by gene number) as an SIV for a habitat and each habitat has $m$ SIVs (arity $m$ as \textit{Habitat}  $H \in SIV^{m}$). For example, if a habitat is $Habitat_{1} = \{12, 345, 26, 7, 141\}$ then the SIVs 12, 345, \ldots, 141 are the selected gene numbers out of say a collection of 500 genes and the subset size is 5 genes. The tuned parameter values for our algorithm are given in the next section. The BBO gene selection algorithm is stated.

\begin{algorithm}
\caption{BBO for Gene Selection}
\label{bbo-gene}
\begin{algorithmic}[1]
 \STATE Initialize BBO parameters
 \STATE Initialize $ecosystem$ with randomly generated $n$ habitats
 \STATE Evaluate habitats : calculate the HSI of each habitat in the $ecosystem$ \textit{(Cross validation accuracies of each gene subset from a classifier)}
 \FOR{$G$ generations}
    \STATE Compute $\lambda_{i}$ and $\mu_{i}$ for each $Habitat_{i}$ based on its HSI
    \STATE Perform $migration$
    \STATE Perform $mutation$
    \STATE Re-evaluate $ecosystem$
    \STATE Perform $elitism$
 \ENDFOR
 \STATE Output the habitat with best HSI and its SIVs (selected genes).
\end{algorithmic}
\end{algorithm}

At each migration or mutation, we ensure that a gene is not duplicated within a single subset of genes, \textit{i.e.} within a single habitat. The subset sizes (the variable $m$) can be set during each run of the BBO algorithm. The ecosystem then has habitats all with same number of SIVs. This selection of subset size is predecided and can be tuned manually by running BBO for various subset sizes.


$Migration$: The migration procedure of the original BBO algorithm \cite{simonbbo} is retained. We produce the algorithm here for the purpose of completeness. 
\begin{algorithm}
\caption{Migration}
\label{bbo-migration}
\begin{algorithmic}[1]
 \STATE Select $H_{i}$ with probability $\propto$ $\lambda_{i}$ 
 \IF{$H_{i}$ is selected}
    \FOR{$j=0$ \TO $n$}
    \STATE Select $H_{j}$ with probability $\propto$ $\mu_{i}$ 
    \IF{ $H_{j}$ is selected}
      \STATE Randomly select an SIV $\sigma$ from $H_{j}$
      \STATE Replace a random SIV in $H_{i}$ with $\sigma$
    \ENDIF
    \ENDFOR
 \ENDIF
\end{algorithmic}
\end{algorithm}

\textit{Mutation with Information Gain based Gene Ranking}: Given the vast search space formed by the possible genes, we keep the mutation rate to about 0.4 to 0.55 in order to graze other portions of this space with a good chance. We have used information gain heuristics as the additional information during mutation. The \textit{information gain} (IG) \cite{info-gain} of a gene is a measure of attribute selection. It stores the \textquoteleft information content\textquoteright \space of a gene with respect to the problem under consideration. The IG of a gene indicates the capability to separate instances for binary classification. We are specially interested in the non-zero IG values. Thus we partition the informative and non-informative genes into separate sets. For IG computation, we have used Weka \cite{weka} data mining software suite, which outputs the information gain based ranking of genes. This is fed to BBO for further computations. This informative gene set with non-zero infogain values are introduced in the population during the process of mutation. 

While in mutation, our algorithm either randomly explores newer genes or exploits from the available set of genes with non-zero infogain values. We set a user defined exploitation probability $q_{0}$. This exploitation is done in a probabilistic manner as analogous with the exploration and exploitation in ant colony optimization \cite{aco-dorigo, aco-specialsection, aco-blum, vkj-aco, vkj-aco2}. To give an example, we have a total of  30 expressed genes and only the first 8 out of these 30 genes have a non-zero infogain. Assuming rand \textless \space $q_{0}$ is satisfied (step 5 of Algorithm 3), then we  select one out of these 8 genes, with a probability proportional to the information gain ranking scores, to be newly put in $Habitat_{i}$ in place of an existing one.  While on the other hand if rand \textless \space $q_{0}$ did not get satisfied, we execute the \textbf{\textit{else}} part (step 7 and 8) $i.e.$ randomly select a gene from all of the 30 genes expressed in the data.

Algorithm 3 gives the detailed mutation algorithm.

\begin{algorithm}[h]
\caption{Mutation}
\label{bbo-mutation}
\begin{algorithmic}[1]
 \FOR{$j=0$ \TO $m$}
    \STATE Use $\lambda_{i}$, $\mu_{i}$ of habitat $H_{i}$ to compute the probability $P_{i}$
    \STATE Select SIV(gene) $H_{i}(j)$ with probability $\propto$ $P_{i}$
    \IF{ $H_{i}(j)$ is selected}
      \IF{ rand \textless \space $q_{0}$  }
	  \STATE \textit{Exploit} : Replace $H_{i}(j)$ with a probabilistically selected a SIV (gene) from the rest (using their information gain)
      \ELSE
	  \STATE \textit{Explore} : Replace $H_{i}(j)$ with a random SIV (gene) out of the rest
      \ENDIF
    \ENDIF
    \ENDFOR
\end{algorithmic}
\end{algorithm}

$Elitism$: We implement elitism so that the best solutions obtained until a particular generation do not get corrupted.

\subsection{Support Vector Machines}
Support Vector Machines (SVMs) \cite{svm} were originally introduced by Vapnik and co-workers \cite{boser} and successively extended by a number of other researchers. SVM employs a maximum margin linear hyperplane for solving binary linear classification problems. For non-linearly separable problems, SVM first transforms the data into a higher dimensional feature and subsequently employs a linear hyperplane. To deal with computational intractability issues it further uses appropriate kernel functions facilitating all computations in the input space itself. Vapnik et al. in \cite{geneVapnik} have themselves used SVM with recursive feature elimination (RFE) for gene selection and achieved notably high accuracy levels. We discuss more about results in the subsequent section.

For our purposes we employ the libSVM \cite{libsvm} library for evaluation of our candidate solutions during each generation.


\subsection{Random Forests}

Random Forests (RF) were first introduced by Breimen and Cutler \cite{rfbreiman}. It is an ensemble of randomly constructed independent decision trees. It performs substantially better than single-tree classifiers such as CART \cite{CART} and C4.5 \cite{c45}. A random subset of attributes are used for node splitting while growing each decision tree. Normally, for each tree, a bootstrap set (with replacement) is drawn from the original training data, $i.e.$ an instance is picked from the training data and is replaced again before drawing the next instance. Likewise, $n$ such instances are taken to form \textquoteleft in bag\textquoteright \space set for a particular tree. For each of the bootstrap training sets, about one \textendash \space third of the samples, on an average, are unused for making the \textquoteleft in bag\textquoteright \space data and are called the \textquoteleft out of bag\textquoteright \space (OOB) data for that particular tree.
The classification tree is built with this \textquoteleft in bag\textquoteright \space data using the CART algorithm \cite{CART}. Separate test data is not required in RF for checking the overall accuracy of the forest. The OOB data is used for cross validation. When all the trees are grown, the $k^{th}$ tree classifies the samples that are OOB
for that tree (left out by the $k^{th}$ tree). In this manner, each instance is classified by about one third of the trees. A majority vote is then taken to decide on the class label for each case. The percentage of times that the voted class label is not equal to the original class of a sample, averaged over all the cases in the training data, is called as the OOB error rate \cite{bagging}.

We have used the randomForest package in $R$ for implementation purposes \cite{RFmanual}.

\section{Discussion and Results}
\subsection{Datasets}
The output of microarray experiments are the expression levels of different genes. Three such datasets were obtained from the Kent Ridge Biomedical datasets repository\cite{datasets} and libSVM repository \cite{libsvm} (made available from various other original sources). 

The Colon Cancer dataset retrieved from the Kent Ridge Biomedical dataset repository consists of 62 instances representing cell samples taken from colon cancer patients. Among these, 40 are tumor samples while 22 otherwise \cite{colon-alon}. The breast cancer dataset is retrieved from the DUKE Breast Cancer SPORE frozen tissue bank \cite{breast-west}. Of the 44 samples we worked with, each sample with expressions for 7129 genes, 22 belong to class A (estrogen receptor-positive ER$+$) while 22 belong to class B (estrogen receptor-negative ER$-$). The Leukemia dataset \cite{leuk-golub} also retrieved from the Kent Ridge Biomedical dataset repository contains the expression of 7129 genes. These are total 72 samples taken from leukemia patients out of which 25 belong to the Acute Myeloid Leukemia (AML) class and 47 belong to the Acute Lymphoblastic Leukemia (ALL) class. These specifications are tabulated in Table I.

\begin{table}[h]
\caption{Dataset Specifications}
\label{dataset_info}
\centering
\begin{tabular}[h]{lccr}

\hline
 Cancer dataset &  \#genes &  \#classes& \#instances\\
 name (D)  &          &           &  (\#A \& \#B)\\
\hline
 Colon (C)  &  2000&  2&  62 (40 \& 22)\\
 Breast (B) &  7129&  2&  44 (22 \& 22)\\
 Leukemia (L)&  7129&  2& 72 (25 \& 47)\\
\hline
\end{tabular}
\end{table}

\subsection{Discussion and Results}

As discussed earlier while describing our algorithms, we have implemented BBO with and without heuristics. Very interestingly, both simple BBO and BBO with heuristics are successful in selecting a good set of features providing comparable classification results. As compared to very typical implementations of gene selection using genetic algorithms and other EAs like PSO \cite{pso-eberhart, ea-de}, which usually work over many generations with large population sizes \cite{compare-evo1}, our implementation of both versions of BBO (that with SVM and RF) for gene selection started showing better results very early with just 40-50 habitats in the ecosystem. We performed 50 simulations each with \#generations varying from 15 to 40. The algorithms almost always converged to comparable results by the end of 25 generations with very minute differences in further generations.

%


\begin{figure}[h]
\centering
\includegraphics[width=2.8in, height=2.5in]{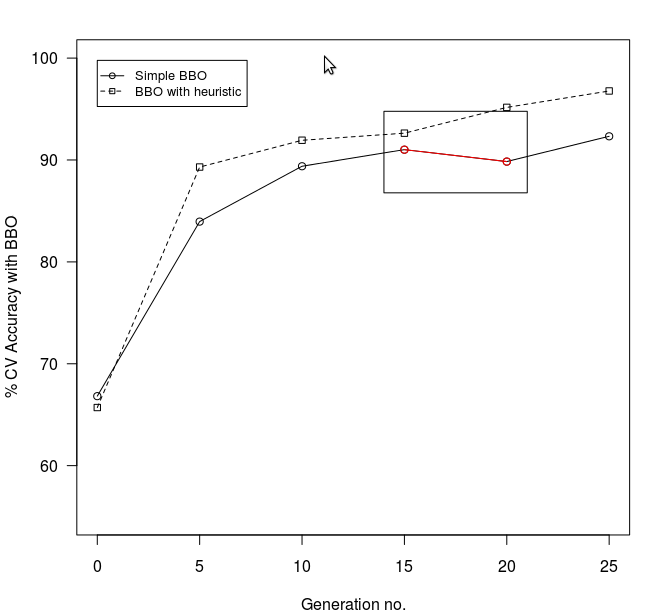}
\caption{Example convergence of population averages of simple vs. heuristic BBO over 25 generations; also with example non-monotonic behavior (boxed) of average suitability of habitats with simple BBO}
\label{fig_graph}
\end{figure}

For BBO with heuristics, we decided not to hinder the original BBO procedure a lot and incorporated the heuristics of information gain of each gene only during the mutation process; even this with only some degree and not for every mutation. This was done by roping in the analogy of exploration and exploitation from the ant colony optimization \cite{aco-dorigo, aco-specialsection, aco-blum} method to give the vast number of other genes a fair chance of inclusion; this degree of exploitation to be controlled by the user depending upon the problem at hand. As a result of this, we observed BBO-SVM and BBO-RF both to converge to an optimal or near optimal solution faster than their counterparts without heuristics. Also, the average suitability of habitats in the ecosystem almost always shows a monotonic improvement in this case unlike the earlier. In effect, it is more like the overall ecosystem (population) showing improvement. The inclusion of probabilistic selection of genes based on entropy, during mutation, adds to the improvement of the overall results. This behavior was consistently observed over 50 simulations. Figure \ref{fig_graph} shows an example of how the population average in BBO with information gain heuristics converged to higher accuracy in lesser generations as compared to simple BBO. The boxed portion demonstrates the typically observed non-monotonic behavior during some runs of simple BBO as against heuristic BBO.

With regards to the classifiers \textendash \space SVM and RF \textendash \space that we have used to evaluate BBO selected gene subsets, there has been work in literature that has reported to find SVM to outperform RF \cite{svm-rf-compare}, both on average and in majority of microarray datasets; and our results reiterate the same. We have verified this observation by training SVM and RF on the same subset of selected genes. The evaluation of each habitat to obtain its HSI from the classifier (the CV accuracies) can be run in parallel resulting in faster results by speeding up the whole process.

From literature, SVMRFE-RG \cite{geneVapnik} and Fisher-RG-SVMRFE \cite{azadeh} report high accuracies with SVM for classification. But the SVMRFE-RG is unable to tackle redundant genes \cite{azadeh}. While in \cite{azadeh}, the authors have used gene ontology to tackle redundant genes during gene selection, we have used the information gain based gene ranking. There are also other methods as in \cite{svm-furey} who have attempted classification of cancer tissue samples with SVMs without feature (gene) selection. They have reported approximately 85\% accuracy of classification (about 10-11 falsely classified samples out of nearly 70 in AML/ALL leukemia cancer case). In \cite{aco-rf}, the authors have used Ant Colony Optimization (ACO) for gene selection with Ant Miner (AM) and RF for classification. 

%
The parameters and their corresponding tuned values used in our algorithms have been listed in Table II. These values were observed to give the most optimum results over extensive simulations.

\begin{table}[h]
\caption{Tuned algorithm parameters}
\label{table2_tuned_param}
\centering

\begin{tabular}[h]{l | r}
\multicolumn{2}{c}{$a$. For BBO with both SVM and RF} \\
\hline
Parameter&  Values  \\
\hline
 Population size (\#candidate solutions for each generation)&  50\\ 
 \#Generations						    &  25\\
 Mutation probability					    &  0.70\\
 Habitat modification probability		 	    &  1.00\\
 Exploitation probability during mutation (for heuristics)  &  0.55\\

\hline
\end{tabular}

\vspace{10pt}
\begin{tabular}[h]{l | r}
\multicolumn{2}{c}{$b$. For SVM} \\
\hline
 Parameter&  Values  \\
\hline
 Cost 	&  50\\ 
 Gamma (for Radial Basis Function as kernel)&  0.02\\
 Folds for cross-validation	& 10\\
 
\hline
\end{tabular}

\vspace{10pt}
\begin{tabular}[h]{l | r}
\multicolumn{2}{c}{$c$. For RF} \\
\hline
 Parameter&  Values  \\
\hline
 Trees in the forest&  500\\ 
 Features per tree&  $\sqrt{\scriptstyle features\_selected\_and\_fed\_to\_RF}$\\
\hline
\end{tabular}

\end{table}

\vspace{5pt}
$Results$ : Table III lists the sizes of gene subsets selected by BBO separately run with SVM and RF algorithms and the 10 \textendash \space fold cross validation accuracies obtained for the selected gene subsets.

\begin{table}[h]
\caption{Subset sizes and best 10-fold cross validation accuracies (CVAs) in \%}
\label{table3_size_accuracy}
\centering
\begin{tabular}[h]{l  c  c  c  c c}
\hline

 D&  Original& 	\#genes &	10-fold & 	 \#genes&	10-fold\\
  &   \#genes&	selected&	CVA for&	selected&	CVA for\\
  & 	   &     BBO-SVM& 	BBO-SVM&	  BBO-RF&	BBO-RF\\
\hline
 C&  2000&  	09&  		98.39&  	{11}&		92.34\\
 B&  7129& 	15&  		99.56&  	{20}&		94.38\\
 L&  7129&  	19&  		99.60&  	{20}&		93.20\\

\hline
\end{tabular}
\end{table}

With reference to literature, BBO-SVM and BBO-RF have fared well as compared to the previously best performing algorithms (for the same colon cancer dataset) namely SVMRFE-RG \cite{geneVapnik}, Fisher-RG-SVMRFE \cite{azadeh}, ACO-AM (Ant Colony Optimization\textendash Ant Miner) and ACO-RF \cite{aco-rf} which had demonstrated accuracies of 93.3, 94.7, 95.47 and 96.77\% respectively \cite{subha,liu}. Similarly, the best performing algorithms for leukemia cancer classification have shown accuracies in the range 91\textendash 97\%, with the best being 97.06\% \cite{aco-rf,geneVapnik,azadeh,roughsets,cong}. While \cite{aco-rf, geneVapnik} and \cite{azadeh} have worked with the same dataset for AML/ALL classification, \cite{roughsets} and \cite{cong} have worked with a different dataset (for Diffuse Large B-Cell Lymphoma (DLBCL)) but with similar properties, which makes us believe that our proposed algorithm with gene selection will also perform equally well as in AML/ALL classification. In case of breast cancer, the reported accuracies, for the dataset we have worked with, have been in the range of 91\textendash 94\%\cite{angela, leuk-golub}. Very clearly, our method of using BBO for gene selection in combination with SVM and RF has outperformed them with accuracies as shown in Table \ref{table3_size_accuracy}. It is worth to note that the methods in literature have almost always reported their best accuracies. In our work, we have reported the average accuracies for both, BBO\textendash SVM and BBO\textendash RF.

\section{Conclusion}

The hybrid BBO-SVM and BBO-RF techniques have shown consistently good results when compared against the highest accuracies for colon cancer, breast cancer and leukemia cancer datasets. Like other evolutionary algorithms, they are also simple to implement, robust and flexible since we can have various possible alternatives as suited to the problem and domain constraints. A significant speedup in the algorithm may be achieved by parallel implementations where the classification accuracies for individual candidate solutions may be computed in parallel. 

\section{Future Work}

Like all EAs, our hybrid methods spur many possibilities of future work \textendash \space some problem dependent and some from the implementation perspective. From problem representation to specific migration and mutation strategies, we can have a variety of schemes. For example, with respect to problem representation, the other suitable scheme that one could explore for BBO here is: each habitat of the ecosystem could have an arbitrary number of attributes (SIVs) which remains fixed for itself across generations but may differ from other habitats in the ecosystem. This is similar to the variable population sizes proposed earlier for BBO  \cite{simonbbo} and other EAs \cite{var-pop-size-ga} but at a finer granularity, variable solution (habitats or chromosomes as the case may be) sizes. This could lead us to a compact implementation framework that can simultaneously output the better performing subset sizes along with the selected genes. Many such variations possible with other EAs could work here too.


\section*{Acknowledgment}

The authors acknowledge the Centre for Modeling and Simulation, University of Pune, India for its support. Also, VKJ gratefully acknowledges the Department of Science and Technology (DST), New Delhi, India for financial support.


\end{document}